# Empirical Analysis of Lifelog Data using Optimal Feature Selection based Unsupervised Logistic Regression (OFS-ULR) Model with Spark Streaming

Sadhana Tiwari[1]
*Department of Information Technology*
*IIIT Allahabad*
*Prayagraj, India*
[1]*rsi2018507@iiita.ac.in*

Sonali Agarwal[2]
*Department of Information Technology*
*IIIT Allahabad*
*Prayagraj, India*
[2]*sonali@iiita.ac.in*

***Abstract-*** Recent advancement in the field of pervasive healthcare monitoring systems causes the generation of a huge amount of lifelog data in real-time. Chronic diseases are one of the most serious health challenges in developing and developed countries. According to WHO, this accounts for 73% of all deaths and 60% of the global burden of diseases. Chronic disease classification models are now harnessing the potential of lifelog data to explore better healthcare practices. This paper is to construct an optimal feature selection-based unsupervised logistic regression model (OFS-ULR) to classify chronic diseases. Since lifelog data analysis is crucial due to its sensitive nature; thus the conventional classification models show limited performance. Therefore, designing new classifiers for the classification of chronic diseases using lifelog data is the need of the age. The vital part of building a good model depends on pre-processing of the dataset, identifying important features, and then training a learning algorithm with suitable hyper parameters for better performance. The proposed approach improves the performance of existing methods using a series of steps such as (i) removing redundant or invalid instances, (ii) making the data labelled using clustering and partitioning the data into classes, (iii) identifying the suitable subset of features by applying either some domain knowledge or selection algorithm, (iv) hyper parameter tuning for models to get best results, and (v) performance evaluation using Spark streaming environment. For this purpose, two-time series datasets are used in the experiment to compute the accuracy, recall, precision, and f1-score. The experimental analysis proves the suitability of the proposed approach as compared to the conventional classifiers and our newly constructed model achieved highest accuracy and reduced training complexity among all among all.

## I. INTRODUCTION

Chronic diseases are the most common and rapidly increasing concern over the world. It causes more deaths than infectious diseases and is one of the significant contributors to the mortality rate [1][2]. Interpretation of the chronic disease data for disease diagnosis is a growing area of research. Before diagnosing the difficult disease, a doctor (physician) has to analyze various factors. Nowadays, techniques like machine learning can be used to fasten up these processes and thus helps in improving the treatment [3][4][5]. Availability of good datasets along with efficient cleaning and pre-processing techniques are essential to develop good classification models for various chronic diseases such as diabetes, heart risks, etc. To make the model more powerful, the best features are selected from the given dataset, so that accuracy of the model gets improved [6]. Many existing techniques, likewise correlation analysis, mutual information between features, recursive eliminated filtering (REF), principal component analysis (PCA), etc. can be used to obtain the most adequate set of features and reduce the dimensionality of the data, to improve the overall performance of the model [7][8].

A massive amount of lifelog data needs to be analyzed for pervasive healthcare monitoring and chronic disease management such as diabetes, heart failure etc. Various big data analytics tools and data mining techniques may help to understand the hidden features of the data pattern and draw some significant outcomes [9]. Clustering is one of the powerful techniques through which any data can be divided into groups [10]. Selecting the appropriate method for making clusters from the dataset and choosing the adequate number of clusters may make the unlabeled as labelled [5]. Multiple clustering methods exist to partition the dataset into groups or classes and handle the outliers present in the data [11] [12]. Handling massive lifelog data of healthcare is very challenging. So, to deal with this challenge, many streaming tools are available; Apache Spark is one of the powerful tools for big data as well as stream data processing which can be easily integrated with all the machine learning models [8] [9] [13].

Many well-established machine learning methods like Logistic Regression (LR) [7], K-Nearest Neighbors (KNN) [14], Naïve Bayes [6][15][16], Decision Trees [6][12][17], Support Vector Machine (SVM) [17][18][19][20], Random Forest (RF) [6][12][20], Artificial Neural Network (ANN) [18][21][22], and deep learning methods [23] such as Long Short-Term Memory (LSTM), Convolutional Neural Network (CNN) [24] etc. are already available. But sometimes they show limited performance due to a lack of efficient data pre-processing. The major offerings of the presented research are:

- To design a new, improved classification model by applying machine learning (ML) and using the

most optimal set of features along with the best set of parameters to achieve more accurate results for effective and reliable clinical outcomes.

- This new hybrid model (OFS-ULR) achieves highest accuracy, less classification time and reduced complexity, which helps in accurate and robust identification of disease from lifelog datasets.
- This research also utilizes the spark streaming environment with machine learning models to evaluate the performance of the diabetes chronic disease data management and classification.
- The evaluation metrics used to evaluate these classification models are accuracy, precision, recall, the area under curve, f1-score, complexity, etc.

The paper is organized as follows: Section 1 describes the introduction part; section 2 discusses the work done in the literature. The proposed methodology has been discussed in section 3. Section 4 shows a detailed idea about the material and methods used for experimental analysis. The result and performance evaluation are incorporated in section 5. The last section 6 provides the concluding remark of this research work and directions for future work.

## II. RELATED WORK

Classification is one of the most important decision-making tools in medical sciences. Sneha et al., have proposed a machine learning-based approach to select the optimal features for early prediction of diabetes mellitus. The outcome shows that the decision tree algorithm and random forest have the highest specificity of 98.20% and 98.00%, respectively, for the analysis of diabetic data. This research also experimented with Naïve Bayesian with the best accuracy of 82.30%. A generalized way is also suggested for the optimal feature selection from the diabetic data set to improve the classification accuracy [6]. In another research, Jain and Singh, surveyed several feature selection and classification techniques for the diagnosis and prediction of chronic diseases were reported. It is reported that dimensionality reduction improves the performance of machine learning algorithms. A broad analysis of feature selection methods is done, and their pros and cons are discussed. It is found that selecting of features play a significant role in enhancing the accuracy of classification systems [7]. Gupta et al., worked on a machine learning technique to provide better healthcare services with less expensive therapeutically equivalent alternatives. Healthcare may have clinical data, claims data, drugs data, hospital data, etc. In this research, clinical and claims data are considered to study of 11 chronic diseases like kidney disease, osteoporosis, arthritis, etc. ML techniques are used to establish the correlation between 11 chronic diseases and the ICD9 diagnostic code for the early prediction of these chronic diseases [8].

Another paper discussed a new model for emotional state classification using various ML classifiers that achieve an accuracy of 75.38 % through ANN classifier. This study was performed on a newly generated data collected from ECG and GSR sensors [12]. Deepika et. al. has worked on some effective mechanisms of chronic disease prediction based on historical health data. Four methods viz., Naïve Bayes, Decision tree, SVM, and ANN classifiers are implemented to diagnose diabetes and heart disease. It is found that SVM gives the highest accuracy rate of 95.556% in the case of heart disease, and in the case of diabetes, the Naïve Bayes classifier gives the highest accuracy of 73.588% [17]. In the research paper [19][20], SVM with RBF-kernel is used as a classification model for the Pima Indian Diabetic dataset evaluated on ROC score. It uses 10-fold cross-validation in the training set to check the robustness of the model. It uses 200 samples for training and 260 samples for testing the model's performance. An accuracy of 78% was achieved. But no feature selection was done on the dataset before training. A significant research provides a fact that in Bangladesh, chronic diseases account for half of the annual death (51%) and near about half of the load of all other diseases (41%). Healthcare sector in developing countries like India, Bangladesh etc. are now generating, collecting, and storing the huge volume of lifelog data. Predictive big data analytics, machine learning and deep learning techniques provides an efficient way to understand the behavior of various chronic diseases in any country or region [21].

ANN [18][19][22], Deep Neural Network [23] such as LSTM [24] and CNN [24] are widely used these days for machine learning tasks that work on the concept of the biological neurons. A hybrid neural network that includes an artificial neural network and fuzzy neural network was used for classification purposes on the Pima Indian Diabetic dataset. The researchers used 10-fold cross-validation to partition the dataset into training and test data. The training duration was 95 seconds and the accuracy achieved was 87.4%. Five hundred samples were used for training, and 2520 samples were used for testing the performance of the model [25]. In a paper, an MLP classifier is used for the classification, and the PIMA Indian Diabetic dataset is used. The training data and test data size is 550 and 218 instances respectively. The accuracy of 77.5% was achieved [26]. It can be seen from the previous research that there is a significant gap in achieving desired accuracy and performance, which may be due to a lack of efficient pre-processing techniques, feature selection, and hyper parameter tuning. Also, there is much scope to work with unlabeled lifelog data of chronic diseases to improve the model performance and to make the model time efficient.

## III. PROPOSED MODEL

The proposed model optimal feature selection based unsupervised logistic regression (OFS-ULR) uses lifelog data such as the United States Chronic Disease Indicator (US-CDI) dataset [27] and diabetes [28], for the experiment. This method is a hybrid technique, including k-means clustering [5][29], PCA [30], and tuned LR model. In our work, k-means clustering is used for associating the class labels to the data, PCA is used for optimal feature selection, grid search method for best hyperparameter selection for LR classifier to improve the performance of US-CDI and UCI-Diabetes dataset. Performance of this hybrid model is compared with existing well-established classification models such as DT, SVM, RF, and GB is also evaluated in the integration of spark streaming environment using both the dataset. The proposed approach used in this work includes the following major steps:

**Step I- Data Cleaning:** Data cleaning helps to clean and refine the dataset by handling missing, duplicates, noise, and outliers in the data. It includes dropping the rows with null values and dropping the features that contain many null values.

**Step II- Data Preparation:** This process makes the data suitable for analysis purposes. It includes steps like feature selection and labelling (if the dataset is unlabeled) using the clustering technique.

**Step III- Normalizing data:** It includes scaling the values of the features to a given scale fit for training.

**Step IV- Training the model:** It includes training different machine learning models on the prepared dataset, evaluating the models on various evaluation metrics, and finding the best model for the classification task. Figure 1 shows the flow diagram of the proposed model.

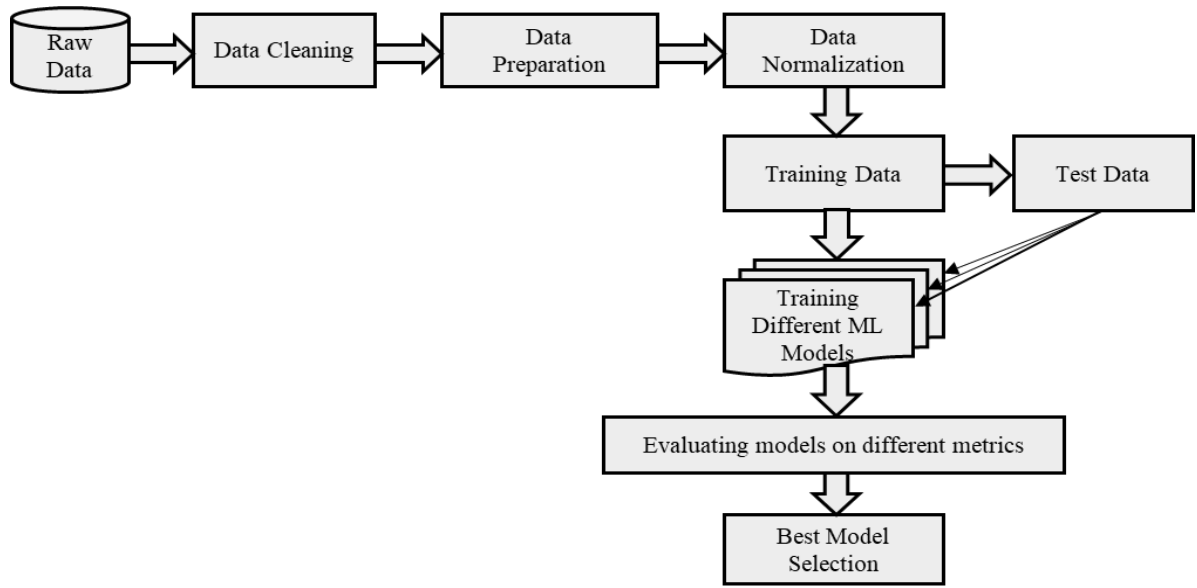

Fig 1. Flow diagram of the proposed model

Algorithm 1 describes the pseudocode of the proposed OFS-ULR model for accuracy computation using the US-CDI and Diabetes dataset. The functionality of the proposed model is divided into three phases. The first phase utilizes a k- means clustering algorithm to handle the problem of unlabeled data. The second phase computes principal components by selecting an optimal feature from the dataset. The third phase handles the overfitting and under fitting issues by selecting the fine-tuned set of parameters for the classification and better model performance.

ALGORITHM I: PSEUDOCODE OF OFS-ULR MODEL

**Inputs:**
D - {d₁, d₂......, dₙ} // Set of n - data instances and can be represented in form of matrix XxY with dimension mxn.
C - {c₁, c₂........cn} // Set of k- cluster centroids;
V− validation data samples
Hyper parameters− (solver, penalty, C)

**Algorithmic Steps:**

**Phase 1:** Associate labels to each data instance using K-means clustering

(i)     Calculate the distance $d(d_i, c_j)$ between all the centroids $(c_{ij})$ and every data instance $(d_i)$
(ii)    Identify the nearest centroid $c_j$ for every data instance and include $d_i$ in-cluster j.
(iii)   Assign cluster_id[i]= j     //j: id of the closest cluster
(iv)    Assign closest_dist[i]= $d(d_i, dj)$
(v)     For every cluster j(1<=j<=k), recompute the centroids;

(vi)      Repeat step (i) to (v)

(vii)     for each data point di

(viii)     If the d ($d_i$, $c_j$) is less than or equal to the current closest cluster, the data
points still stay in the same cluster

(ix)     Else include the data point is some other cluster

(i)     End for

**Phase 2:** Optimal Feature Selection

(i)     Systematize dataset D across column Y to make mean

(ii)     Generate the covariance matrix S from D using the formula

$$COV = \frac{\sum_{i=1}^{n}(X_i - \bar{x})(Y_i - \bar{y})}{n-1} \quad // \bar{x} \text{ and } \bar{y} \text{ denotes the means of X and Y}$$

$$S = \frac{1}{(n-1)}\sum_{i=1}^{n}(X_i - \bar{X})(X_i - \bar{X})' \ // \ \bar{X} \text{ is the mean vector consists the averages of the n variables}$$

(iii)     Decompose the matrix S and compute eigen vectors and eigen values

(iv)     Arrange eigen values in descending manner to rank the correlated eigen vectors

(v)     Choose top N eigenvectors corresponding N largest eigen values, where N is the dimension of    new features subset ($N <= Yn$)

(vi)     Transform the actual dataset D into the new feature subspace using the top N eigen vectors

**Phase 3:** Parameter tuning and classification model setup

(i)     Function GridSearchCV (solver, penalty, C)

(ii)     Apply grid search on hyperparameter

(iii)     return paramgrid hyperparameter estimator

(iv)     Function LogisticRegression ($D\_train$, V$\_train$, $D\_test$, V$\_test$)

(v)     Use param grid for getting the best hyperparameter estimator

(vi)     Fit the model with $D\_train$, V$\_train$

(vii)     Return classification result

**Output:**

Cluster C, Optimal subset of features, Classification result using paramgrid

### IV. MATERIAL AND METHODS

This section discusses the material and methods applied in this work. The first sub-section describes the dataset in detail, including the data preparation phase. The experiment setup and process of experimenting are discussed in subsection two. The third subsection provides a brief idea about the evaluation metrics used for performance evaluation in the presented work.

**4.1 Dataset description**: Two well-known datasets are used for this experiment in the proposed technique, i.e., the US-CDI dataset and the Diabetes dataset.

### I. US-Chronic Disease Indicator dataset:

This dataset is prepared by CDC's Division of Population Health. It provides many indicators developed by consensus and allows states and territories and large metropolitan areas to uniformly define, collect, and report chronic disease data that are important to public health practice and available for states, territories, and large metropolitan areas. It contains 34 attributes and around 237961 instances [27]. Table 2 given below describes the dataset:

Table 2: Attribute description of US-CDI

| Attribute Name | Null | Not Null | Dtypes | %valid Values |
|---|---|---|---|---|
| Year Start | 0 | 237961 | int64 | 100.000000 |
| Year End | 0 | 237961 | int64 | 100.000000 |
| Location Abbr | 0 | 237961 | String, Categorical | 100.000000 |
| Location Desc | 0 | 237961 | String, Categorical | 100.000000 |
| Data Source | 0 | 237961 | String, Categorical | 100.000000 |
| Topic | 0 | 237961 | String, Categorical | 100.000000 |
| Question | 0 | 237961 | String, Categorical | 100.000000 |
| Response | 160105 | 77856 | mixed-type(int/float/str) | 32.717966 |
| Data Value Unit | 32743 | 205218 | String, Categorical | 86.240182 |
| Data Value Type ID | 0 | 237961 | String, Categorical | 100.000000 |
| Data Value Type | 0 | 237961 | String, Categorical | 100.000000 |
| Data Value | 48548 | 189413 | String, Continuous | 79.598338 |
| Data Value Alt | 71965 | 165996 | float64, Continuous | 69.757649 |
| Data Value Footnote Symbol | 109556 | 128405 | String | 53.960523 |
| Data value Footnote | 109739 | 128222 | String | 53.883620 |
| Low Confidence Limit | 96573 | 141388 | float64, Continuous | 59.416459 |
| High Confidence Limit | 96573 | 141388 | float64, Continuous | 59.416459 |
| StratificationCategory1 | 0 | 237961 | String, Categorical | 100.000000 |
| Stratification1 | 0 | 237961 | String, Categorical | 100.000000 |
| StratificationCategory2 | 160105 | 77856 | mixed-type(int/float/str) | 32.717966 |
| Stratification2 | 160105 | 77856 | mixed-type(int/float/str) | 32.717966 |
| StratificationCategory3 | 160105 | 77856 | mixed-type(int/float/str) | 32.717966 |
| Stratification3 | 160105 | 77856 | mixed-type(int/float/str) | 32.717966 |
| Geo Location | 1438 | 236523 | String, Continuous | 99.395699 |
| Topic ID | 0 | 237961 | String, Categorical | 100.000000 |
| Question ID | 0 | 237961 | String, Categorical | 100.000000 |
| Response ID | 160105 | 77856 | mixed-type(int/float/str) | 32.717966 |
| Location ID | 0 | 237961 | mixed- type(int/float/str), Categorical | 100.000000 |
| StratificationCategoryID1 | 0 | 237961 | String, Categorical | 100.000000 |
| StratificationID1 | 0 | 237961 | String, Categorical | 100.000000 |
| StratificationCategoryID2 | 160104 | 77857 | mixed-type(int/float/str) | 32.718387 |
| StratificationID2 | 160104 | 77857 | mixed-type(int/float/str) | 32.718387 |
| StratificationCategoryID3 | 160105 | 77856 | mixed-type(int/float/str) | 32.717966 |
| StratificationID3 | 160105 | 77856 | mixed-type(int/float/str) | 32.717966 |

After applying PCA, optimal subset of features could be obtained for the betterment of the developed model. The plot given in figure 3 shows the number of valid and invalid instances in each of the feature according to the importance of each feature as shown in figure 2.

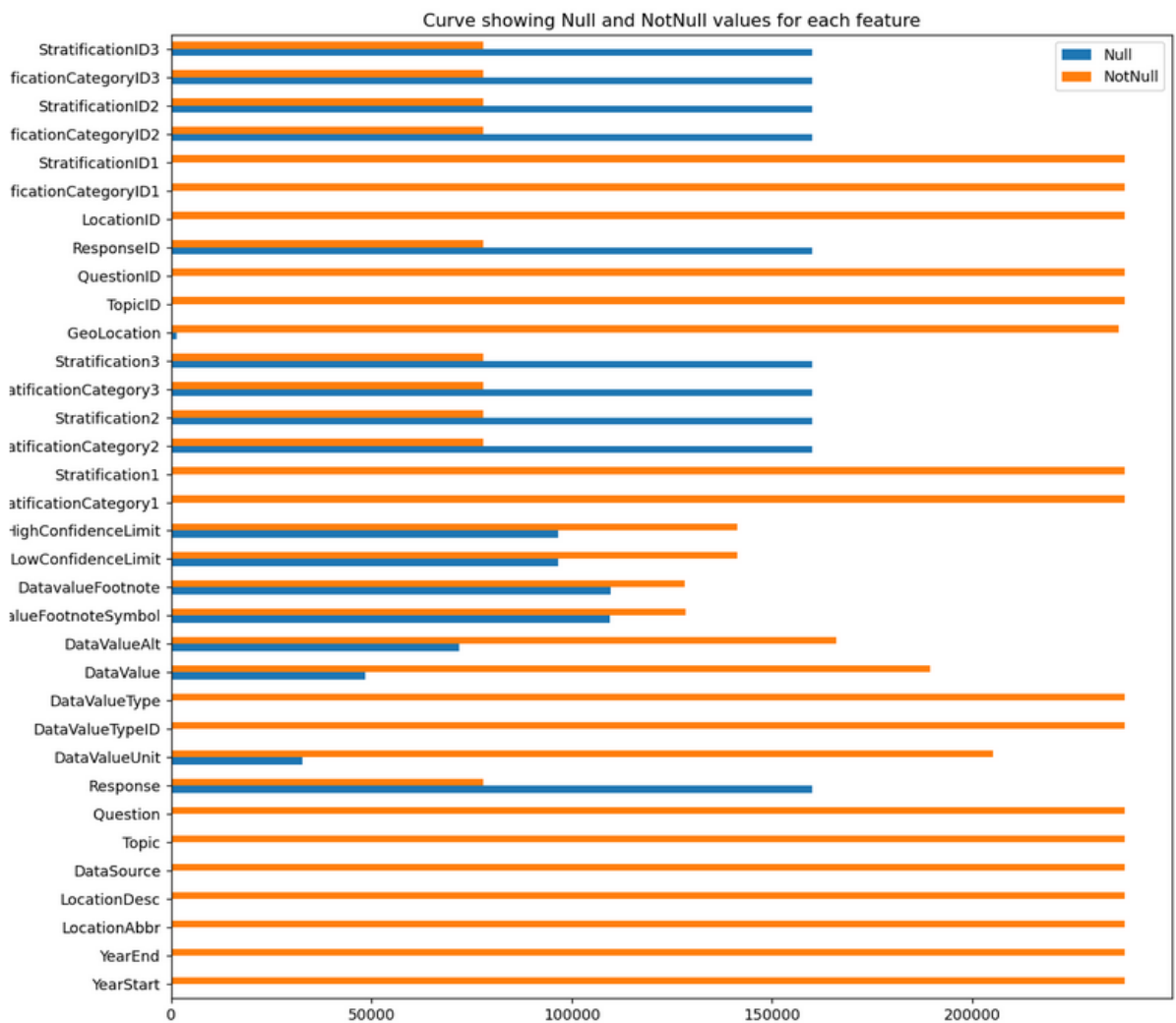

Fig. 2. Valid and invalid features in US-CDI data

**Data Preparation**:

Many features like Location Abbr, Location ID, and Location Desc encode the same information in the dataset. So, after inspecting each feature carefully and dropping the redundant attributes and the attributes with a majority of null values, we get these attributes: Year Start, Year End, Location Id, Data Source, Topic ID, Data Value Unit, Data Value Type ID, Data Value, Low Confidence Limit, High Confidence Limit, Geo Location, Question ID, Geo Location, Stratification ID1, Stratification Category ID1. After performing these feature selections, instances with null values are dropped, and the geolocation feature is decomposed into two features, Geo_lat and Geo_lon, containing the latitude and longitude of the location. After data preparation, 140298 instances were there in the final dataset. Afterwards, Data Value, Low Confidence Limit, High Confidence Limit, Geo_lat and Geo_lon features have been used for applying k-means clustering to associate the class labels, because the data is unlabeled. For selecting optimal number of clusters i.e. 2, elbow test has been used. Basically, the dataset is divided into classes or clusters (figure 3).

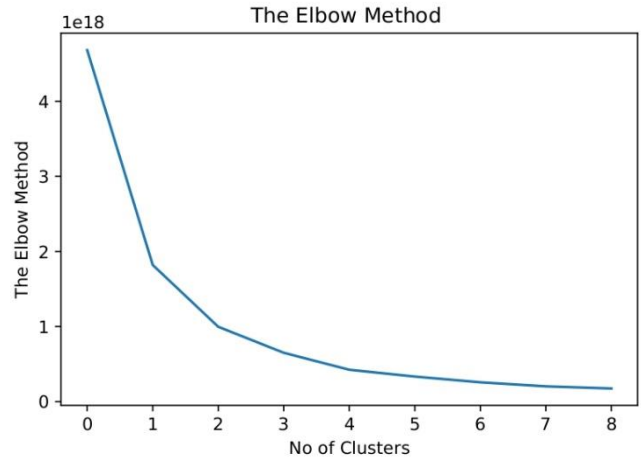

Fig. 3. Optimal no. of cluster using elbow test for US-CDI dataset

**Training and Test Data:**

This dataset is divided into training data and test data, with 80% of data for training and 20% of data for testing the model. So, training data contains 112238 samples while test data contains 28060 samples.

**II. Diabetes Dataset:**

It is a multivariate, time-series dataset on the UCI repository. This dataset contains four fields, per record, namely Date (MM-DD-YYYY format), Time (XX: YY format), Code, and Value, as shown in table 1. The code field is categorical, and the Value field is continuous. It is an unlabeled dataset. It has 29330 instances in total. There are 33 null values in the Date attribute, 154 invalid values in the Code attribute, and 33 invalid values in the Value attribute [28].

Table 1: Diabetes dataset description

| Dataset name | Resource | Attributes | Instances | Classes |
|---|---|---|---|---|
| Diabetes Dataset | UCI-Repository | 4 | 29330 | 2 |

**Data Preparation:**

After dropping all the rows with invalid values, 29143 valid instances are left to work. Also, Date and time features are combined to form a feature named timestamp and converted to posix time. As the dataset is unlabeled, labelling is done by using k-means clustering, and data is divided into two clusters, i.e., classes. To decide the number of clusters from unlabelled Diabetes dataset, the elbow method is used to choose the optimal number of clusters as 2, where clustering performance has been validated with different k values as 2,3,4,5…10 as shown in figure 4.

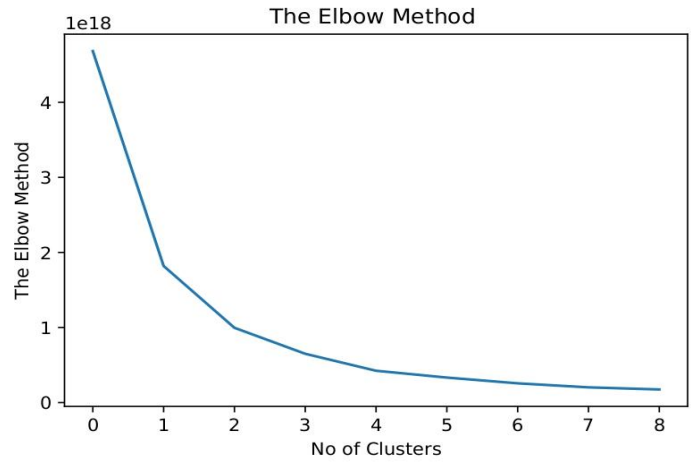

Fig. 4. Optimal no. of cluster using elbow test for UCI- Diabetes dataset

PCA is used to project the feature vector to three-dimension space for plotting it. Then clusters were identified, and labelling is performed using k-means clustering (figure 5).

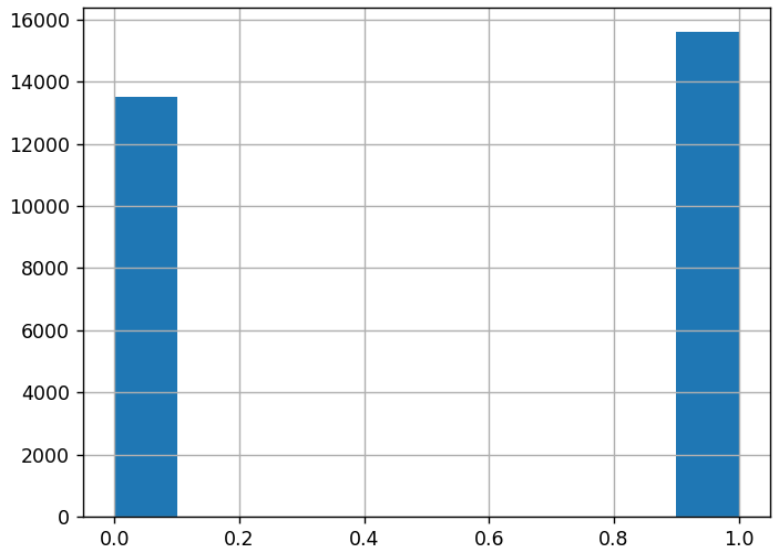

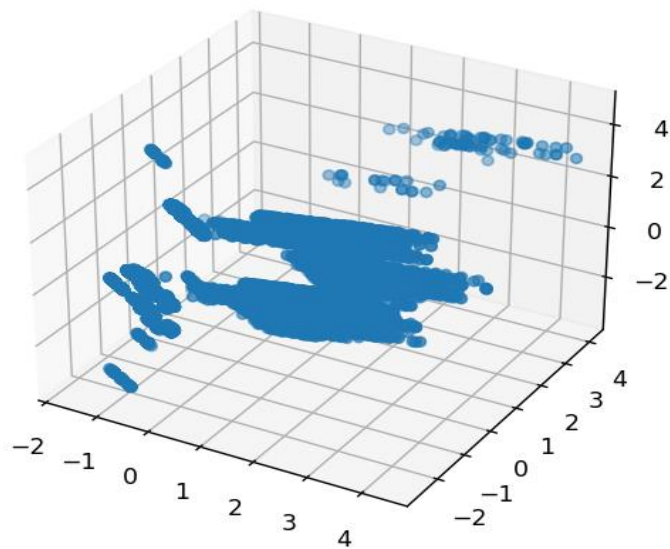

Fig. 5. No. of instances in each class labels in the dataset

**Training and Test Data:**

After labelling the dataset, it is divided into training and test data, with 80% of data for training and 20% of data for testing the model. So, training data contain 23314 samples while test data contains 5828 samples.

### 4.2 Experimental setup

**Execution Environment**

All the models for the execution have been developed and tested for both the chronic disease lifelog dataset. The experiments are conducted on Apache Spark 3.0.1 using python API (pyspark). The datasets are stored as comma-separated values (.csv) files and directly read in the program.

**Training and testing mechanism**

The dataset has been divided into training and test data in the proportion 4:1. Different models are trained on the training data, and for the robust model, 3-fold cross-validation was also performed during the training phase. Then the models are evaluated on the test data to access the performance. Firstly, the models are trained on data with all the features and then evaluated on the test data to assess their performance, and some features were selected based on the domain knowledge and the again the models are trained on the data with selected features to compare the performance difference these two approaches. And lastly, trained models were also evaluated on streams of data. The models are evaluated on different evaluation metrics, and results are compared for each approach.

**Under fitting and over fitting**

Underfitting is a situation where the model does not perform well both on training and test data. At the same time Overfitting is a situation where the model performs well on training data but fails to perform well on the test data. So, to avoid such situations, great care is taken in hyperparameter tuning. The models were tested both on training data and test data to check such situations.

### 4.3 Evaluation Metrics

To analyse the performance of newly designed hybrid classification model, the accuracy, precision, recall, and Receiver Operating Characteristics (ROC) measures are adopted. The following are some basic terminologies:

**True Positive (Tp)**: True Positive can be defined as the number of examples correctly classified to that class.

**True Negative (Tn)**: True Negative are the number of examples correctly rejected from that class.

**False Positive (Fp)**: False Positive denotes the number of examples incorrectly rejected from that class.

**False Negative (Fn)**: False Negative represents the number of examples incorrectly classified to that class.

**Accuracy (Ac)**: It is defined as the percentage of correct predictions for the data.

Accuracy (Ac) = (Tp + Tn) / (Tp + Tn + Fp + Fn)

**Precision (Pr)**: It measures how many of the samples predicted as positive are positive. It is used as a performance metric when the goal is to limit the number of false positives.

Precision (Pr) = Tp / (Tp + Fp)

**Recall (Re)**: It measures how many positive samples are captured by the positive predictions. It is used as a performance metric to identify all positive samples.

Recall (Re) = Tp / (Tp + Fn)

**F-measure (Fm)**: It summarizes the precision and recall score. It is the harmonic mean of precision and recall.

F-measure (Fm) = 2 * (Pr * Re) / (Pr + Re)

**Receiver Operating Characteristics (ROC)**: It reports results in terms of false-positive rate (FPR) and true-positive rate (TPR). ROC curves show the trade-off between FPR and TPR for different thresholds. ROC curve can be summarized by using a single number, i.e., the area under the curve known as AUC.

FPR = Fp / (Fp + Tn)

TPR = Tp / (Tp + Fp)

## 5. Result and Performance Evaluation

The classification experiments are conducted on the US-CDI dataset and UCI-Diabetes dataset using well-established machine learning models such as OFS-ULR, DT, SVM, RF, and GB. All the used classification models are trained on these datasets, summarized in the following tables. Table 3 presents the performance evaluation using all the existing models and the proposed model using US-CDI dataset. Table 4 shows the results of all the performance metrics using UCI-Diabetes dataset.

# I. US-CDI DATASET

Table 3: Performance evaluation using Different classifiers

| Classifier Used | Avg. time of classification | Ac | Fm | Pr | Re | Significance level | Time and Space complexity |
|---|---|---|---|---|---|---|---|
| OFS-ULR | 2min 24s | 0.99960 | 0.99960 | 0.99787 | 1.0 | 0.05 | O(nd) and O(d) |
| SVM | 4min 31s | 0.96243 | 0.96452 | 1.0 | 0.99743 | 0.05 | O(n²) and O(kd) |
| Decision Tree Classifier | 1min 22s | 0.95298 | 0.95454 | 1.0 | 0.95866 | 0.05 | O(n*log(n)*d) and O (max depth of tree) |
| Random Forest Classifier | 3min 26s | 0.98640 | 0.98541 | 0.99941 | 0.99495 | 0.05 | O(n*log(n)*d*k) And O (depth *k) |
| Gradient Boosted Trees Classifier | 3min 12s | 0.99923 | 0.99921 | 1.0 | 0.99912 | 0.05 | O(n*log(n)*d*k) and O(depth*k) |

# II. UCI- DIABETES DATASET

Table 4: Performance evaluation using Different classifiers

| Classifier Used | Avg. time of classification | Ac | Fm | Pr | Re | Significance level | Time and Space complexity |
|---|---|---|---|---|---|---|---|
| OFS-ULR | 42.9s | 0.99982 | 0.99981 | 0.999635 | 1.0 | 0.05 | O(nd) and O(d) |
| SVM | 53.8s | 0.99948 | 0.99945 | 1.0 | 0.99890 | 0.05 | O(n²) and O(kd) |
| Decision Tree Classifier | 18.2s | 0.99931 | 0.99927 | 1.0 | 0.99854 | 0.05 | O(n*log(n)*d) and O (max depth of tree) |
| Random Forest Classifier | 26.9s | 0.99949 | 0.99944 | 0.999257 | 0.99962 | 0.05 | O(n*log(n)*d*k) And O (depth *k) |
| Gradient Boosted Trees Classifier | 33.1s | 0.99931 | 0.99931 | 1.0 | 0.99854 | 0.05 | O(n*log(n)*d*k) and O(depth*k) |

The plots given in figure 6 show accuracy evaluation scores of different classification models trained on the US-CDI dataset and the Diabetes dataset. This plot shows that the logistic regression has the highest accuracy for both datasets.

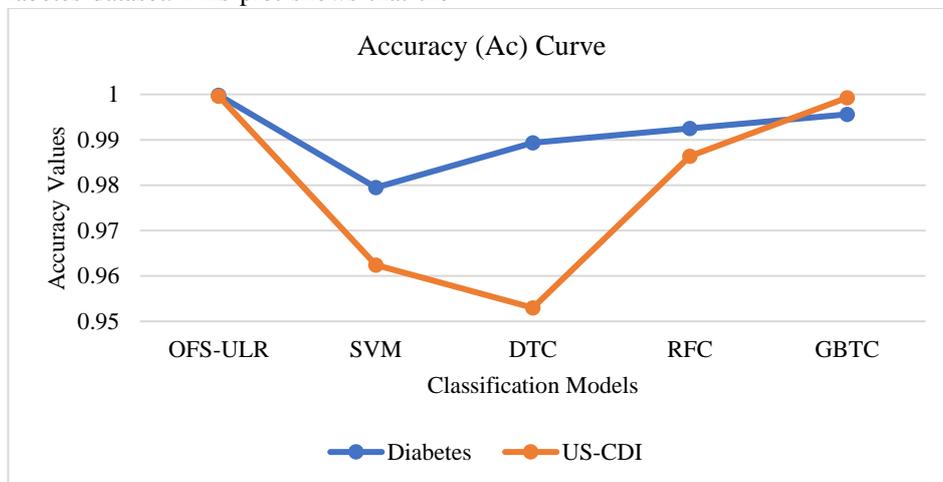

Fig. 6. Accuracy values for different models

The plot mentioned in figure 7 shows the f1-score of different models trained on both the used dataset in the experiment. This plot shows that the Logistic Regression model has the highest f1-score among all models for the datasets used in this work.

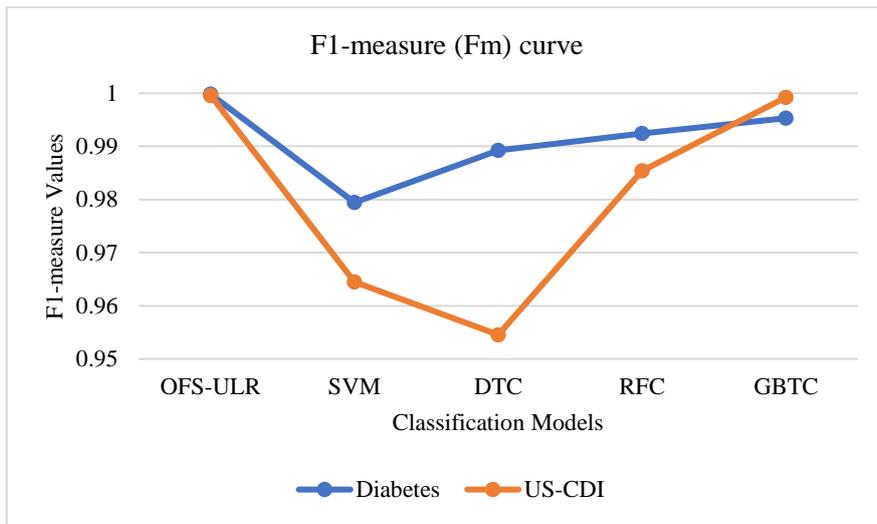

Fig. 7. F1-measure values for different models

The plot attached in figure 8 represents the precision value of different models trained on the US-CDI and the Diabetes dataset. This plot shows SVC has the highest precision among all models for the Diabetes dataset and Decision Tree Classifier has the highest precision for the US-CDI dataset.

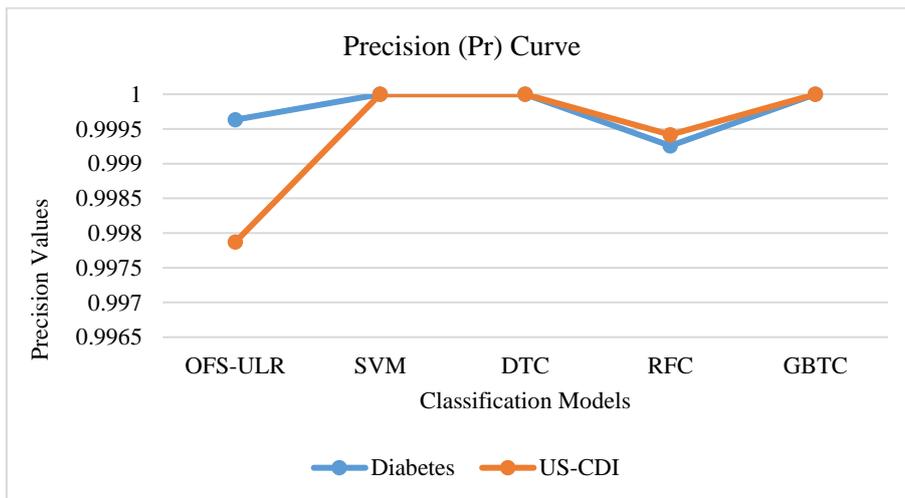

Fig. 8. Precision values for different models

The plot below in figure 9 shows the recall values of different models trained on the US-CDI and the Diabetes dataset. This plot shows the Logistic Regression model gives the highest recall values for both the datasets among all models.

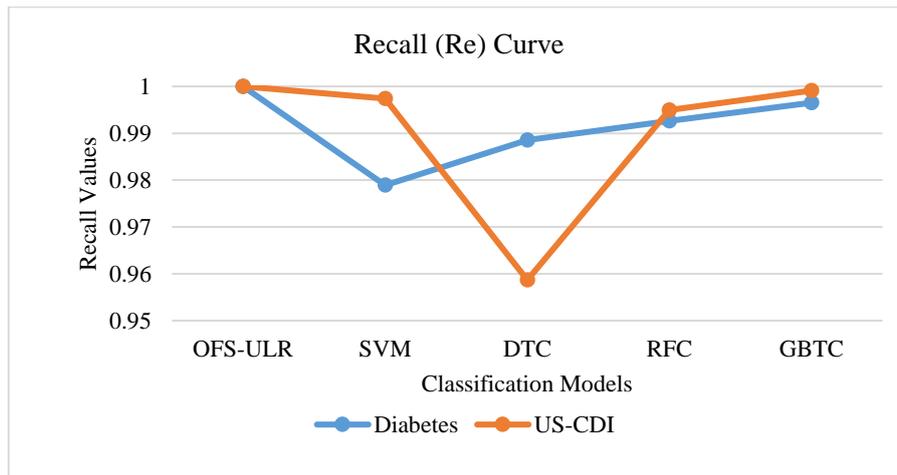

Fig. 9. Recall values for different models

The plot attached in figure 10 shows the time taken by different models for training using the US-CDI and diabetes dataset. This plot shows SVC takes maximum time for training, and Decision Tree Classifier takes minimum time for training.

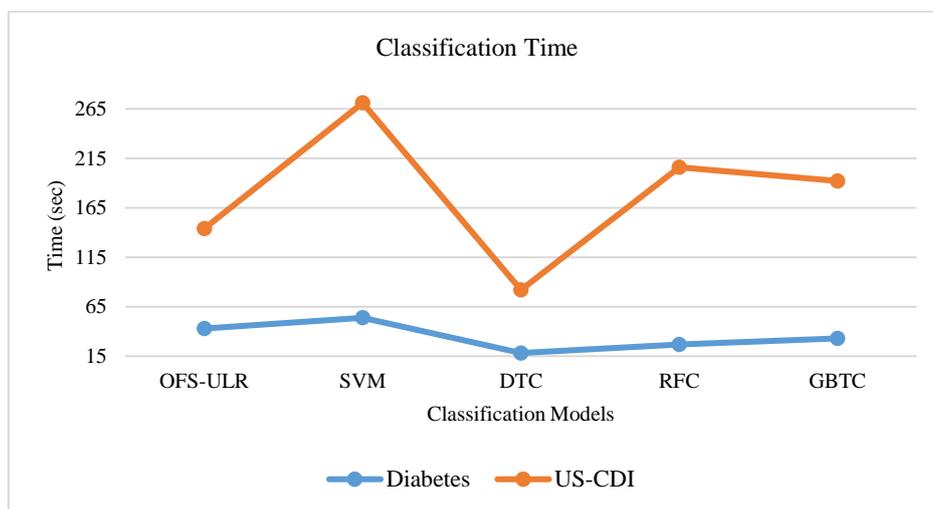

Fig. 10. Time taken by different models for training

## Analysis of Diabetes and US-CDI dataset using Dstreams

The classification experiments are conducted on the US-CDI datasets, and UCI-Diabetes datasets using well-established machine learning models such as OFS-ULR, SVM, DT, RF and GB in spark streaming environments. All the used classification models are trained on these datasets, summarized in the following tables. Table 5 presents the performance evaluation using the existing models and the proposed model using the US-CDI dataset. Table 6 shows the results of all the performance metrics using the Diabetes dataset in a streaming environment. The trained models were also evaluated on streams of test data. The streams of test data were preprocessed, and then the transformed data was used for prediction. Figure 11 shows a neat illustration of our workflow:

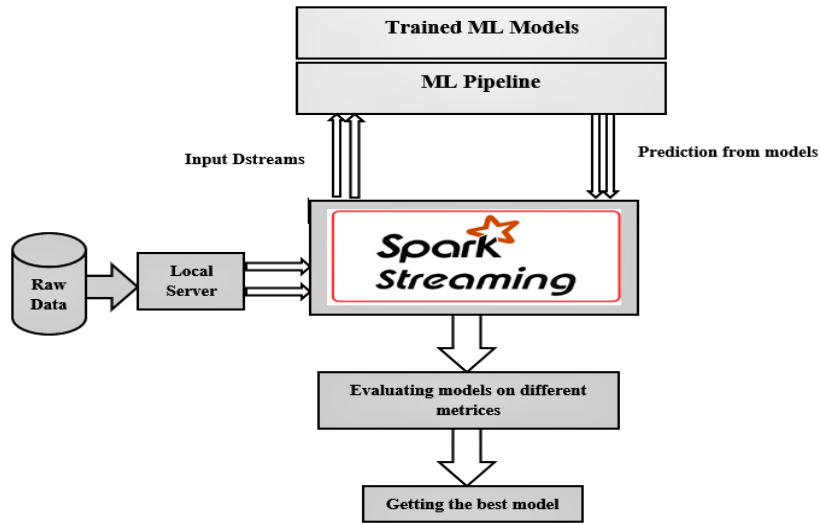

Fig. 11. Workflow of the proposed model using Spark streaming

The evaluation using Dstreams is performed on a sample of test data for US-CDI dataset and the results are summarized in table 5:

Table 5: Performance evaluation using Different classifiers on US-CDI dataset using Dstreams

| Classifier Used | Ac | Fm | Pr | Re |
|---|---|---|---|---|
| OFS-ULR | 0.998502 | 0.985671 | 1.0 | 0.5 |
| SVM | 0.645122 | 0.641652 | 0.5 | 0.5 |
| Decision Tree Classifier | 0.943251 | 0.938672 | 0.5 | 0.5 |
| Random Forest Classifier | 0.976734 | 0.956342 | 1.0 | 0.5 |
| Gradient Boosted Trees Classifier | 0.943312 | 0.946322 | 1.0 | 0.5 |

The evaluation using Dstreams is performed on a sample of test data for diabetes dataset and the results are summarized in table 6:

Table 6: Performance evaluation using Different classifiers on Diabetes dataset using Dstreams

| Classifier Used | Ac | Fm | Pr | Re |
|---|---|---|---|---|
| OFS-ULR | 1.0 | 1.0 | 1.0 | 1.0 |
| SVM | 1.0 | 1.0 | 1.0 | 1.0 |
| Decision Tree Classifier | 1.0 | 1.0 | 1.0 | 1.0 |
| Random Forest Classifier | 1.0 | 1.0 | 1.0 | 1.0 |
| Gradient Boosted Trees Classifier | 1.0 | 1.0 | 1.0 | 1.0 |

Parameters like average time of classification, significance level, and time and space complexity are the same for both techniques, i.e., with/without stream processing. So these parameters are included only once for each of the datasets. So while comparing the performance of the designed and all the existing classifiers in spark streaming environment using Dstreams, it can be concluded that the accuracy will be improved in real-time environment.

The plots in figure 12 shows the accuracy of different classification models trained on the US-CDI and Diabetes dataset using spark streaming environment. Although for diabetes dataset every model gives similar accuracy in the streaming environment, while taking the US-CDI dataset, the OFS-ULR model gives the best performance.

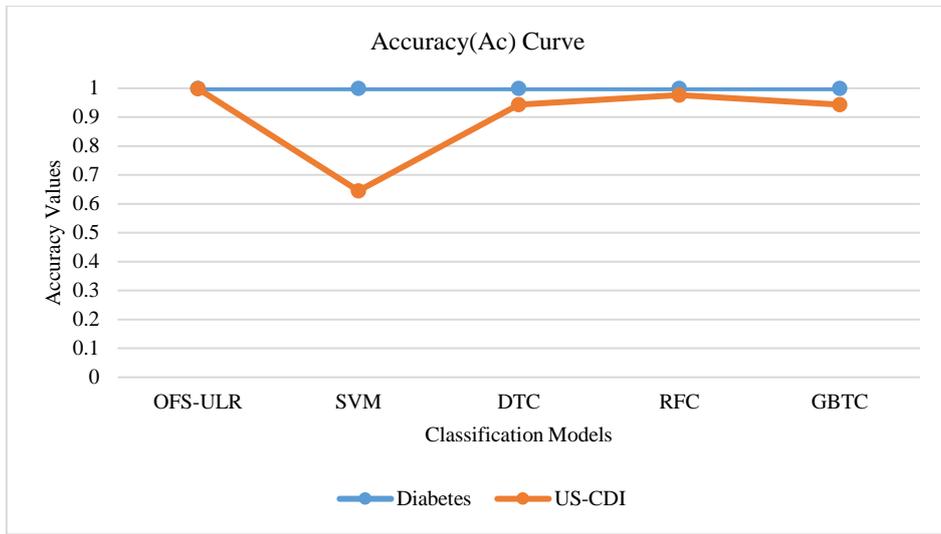

Fig. 12. Accuracy values for different models using Dstreams

The plot mentioned in figure 13 shows the f1-score of different models trained on both the used dataset in the experiment in a spark streaming environment. This plot shows the OFS-ULR model has the highest f1-score among all models for both the datasets used in this work.

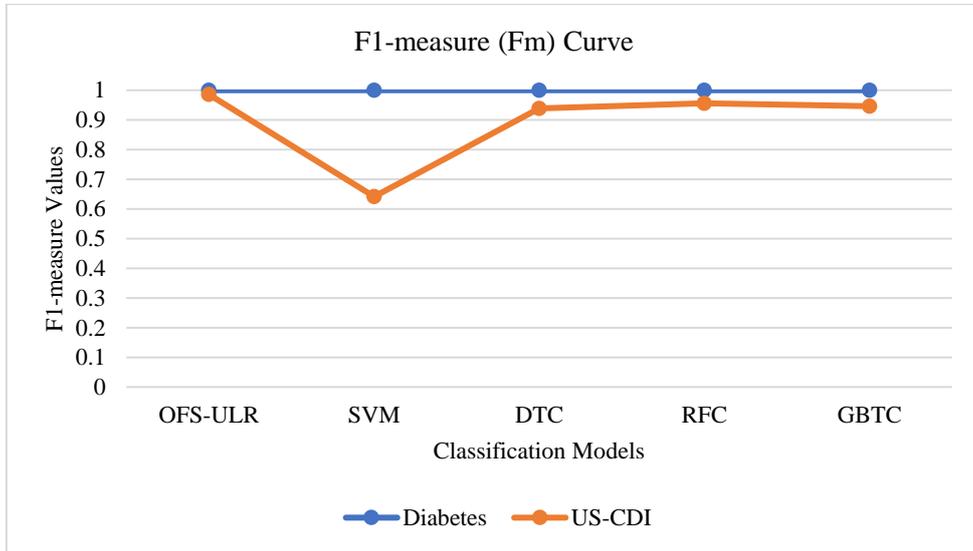

Fig. 13. F1- measure values for different models using Dstreams

The plot attached in figure 14 represents the precision value of different models trained on the US-CDI and Diabetes dataset in a spark streaming environment. This plot shows OFS-ULR, RFC and GBTC have the same precision among all models for both the dataset among all the models.

The plot below in figure 15 shows the recall value of different models trained on the US-CDI and Diabetes dataset using a streaming environment. This plot shows that all the models give the same recall values for both datasets.

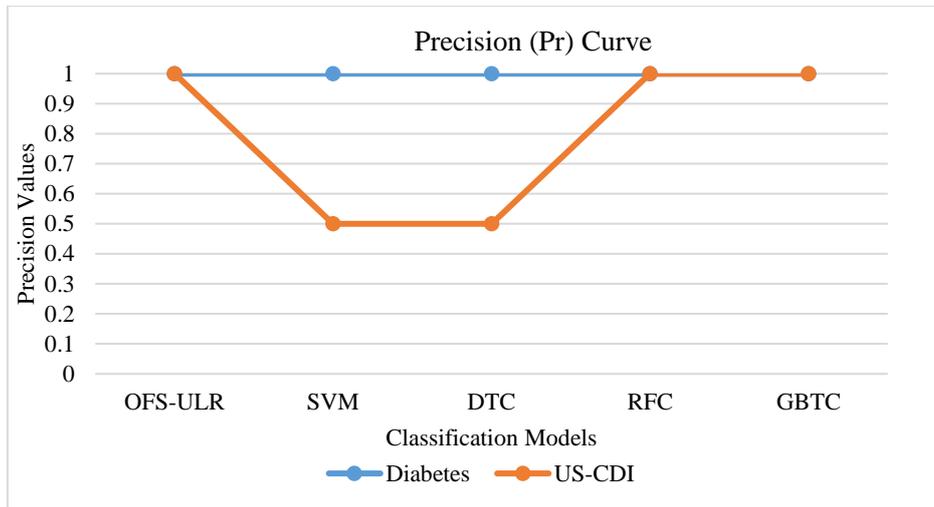

Fig. 14. Precision values for different models using Dstreams

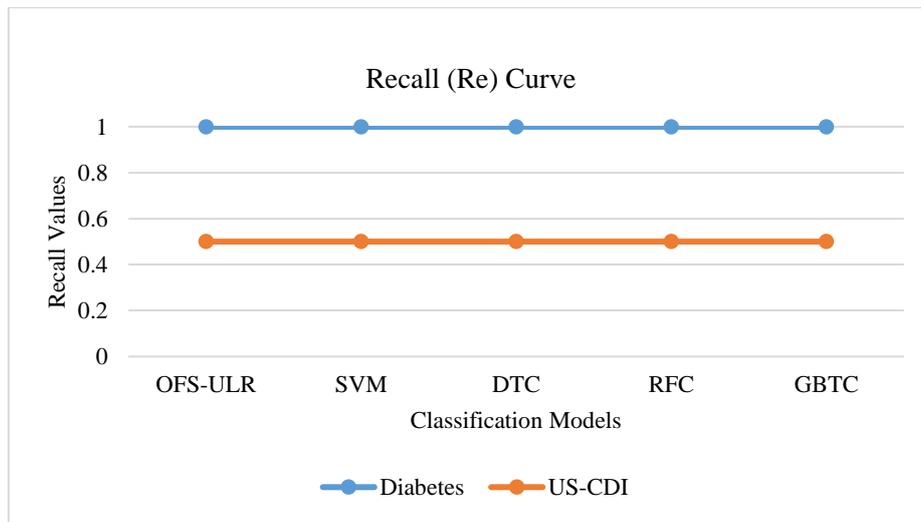

Fig. 15. Recall values for different models using Dstreams

**Comparative Analysis of existing methods**

Table 7 proves that the accuracy achieved using the proposed model is highest among all the previously reviewed models available. It shows a comparison of the methods applied for existing approaches and the proposed approach in this paper:

Table 7: Comparative analysis of the proposed model

| Dataset | Method | Accuracy (%) | Reference |
|---|---|---|---|
| US-CDI Time series dataset | **OFS - ULR** | **99.9%** | Proposed approach |
| | Logistic Regression and Random forest | 99.23% | Jiongming Qin, Lin Chen, Yuhua Liu, Chuanjun Liu, et. al.[25] |
| | SVM | 95.2% | Kumari Deepika, S. Seema [17] |
| | Neural Networks | 86.8% | H. Kahramanli, N. Allahverd [22] |
| | ANN | 75.38% | Tiwari, Sadhana, et al. [12] |
| | Naïve Bayes | 82.30% | N. Sneha and Tarun Gangil [6] |
| | GroupNet | 81.13% | Xiaoqing Zhang, Hongling Zhao, et. al. [26] |
| | MLP | 80% | Saumendra Kumar Mohapatra, et. al. [23] |
| | SVM | 78% | V. Anuja Kumari ,R.Chitra [20] |
| Diabetes Time series dataset | **OFS - ULR** | **99.9%** | Proposed approach |
| | Logistic Regression and Random forest | 99.23% | Jiongming Qin, Lin Chen, Yuhua Liu, Chuanjun Liu, et. al. [25] |
| | SVM | 95.2% | Kumari Deepika, S. Seema [17] |
| | Neural Networks | 86.8% | H. Kahramanli, N. Allahverd [22] |
| | ANN | 75.38% | Tiwari, Sadhana, et al. [12] |
| | Naïve Bayes | 82.30% | N. Sneha and Tarun Gangil [6] |
| | GroupNet | 81.13% | Xiaoqing Zhang, Hongling Zhao, et. al. [22] |

| | MLP | 80% | Saumendra Kumar Mohapatra, et. al. [23] |
|---|---|---|---|
| | SVM | 78% | V. Anuja Kumari ,R.Chitra [20] |

## VI. CONCLUSION AND FUTURE SCOPE

In this paper, two lifelog datasets have been used for chronic disease classification. Since lifelog datasets usually carry missing values and other impurities, thorough cleaning and pre-processing steps were performed to prepare them for further processing. Optimal features get selected using PCA to improve the overall performance of the model. We also performed unsupervised learning to generate the class labels from the lifelog data samples. Later all these data are used to train different ML models, and these models are evaluated by several evaluation metrics such as accuracy, precision, recall, f1-score, and ROC curve. The proposed technique has achieved good results on both datasets. The newly constructed model OFS-ULR performs well for both datasets with an accuracy of 99.9% and 99.8% respectively in static environment. On the other hand, the proposed method also gives good accuracy, i.e., 99.8% for US-CDI data and 100% for diabetes data in dynamic environment using spark streaming. The model reduces the complexity of the algorithm and response delay in a dynamic environment. Hence this classification can be used for detection of chronic diseases such as Diabetes. Further, the designed model performance can be improved by handling imbalance class problem to avoid biased results. In future, the proposed approach can also be used with other techniques to make better models on similar datasets which can facilitate the faster detection and diagnosis of chronic diseases and help to reduce the adverse effect of these diseases on society.

ACKNOWLEDGMENT

The authors are thankful to the Indian Institute of Information Technology, Allahabad (IIITA) for providing all the necessary infrastructure and support.

AUTHOR'S BIOGRAPHY

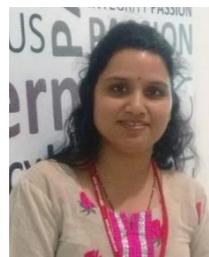 Sadhana Tiwari received the B. Tech. Degree in Computer with a focus in CSE and completed the M. Tech. Degree in Software Engineering from IIIT Allahabad, India. Currently I am working as a research scholar in IIIT Allahabad, India in the area of big data analysis and stream data mining in biomedical domain.

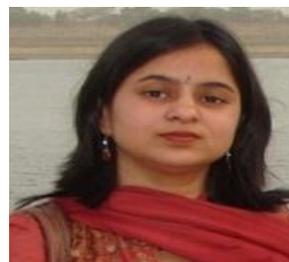 Sonali Agarwal, working as an Associate Professor in the Information Technology Department of the Indian Institute of Information Technology (IIIT), Allahabad, India. She received her Ph. D. Degree at IIIT Allahabad and joined as faculty at IIIT Allahabad since October 2009. The main research interests are in the areas of Stream Analytics, Big Data, Stream Data Mining, Complex Event Processing System, Support Vector Machines and Software Engineering. Dr. Sonali Conducted many workshops and seminars related to the area of big data analytics and real time stream analytics. She received many funded projects form DST, SERB and CSIR etc.